\documentclass[runningheads]{llncs}

\usepackage{times}
\usepackage{soul}
\usepackage{url}
\usepackage[hidelinks]{hyperref}
\usepackage[utf8]{inputenc}
\usepackage{graphicx}
\usepackage[switch]{lineno}

\usepackage{booktabs}
\usepackage{multirow} 
\usepackage{xspace}
\usepackage{amssymb,amsmath}
\usepackage{stmaryrd}
\usepackage{cleveref}
\usepackage{xcolor}
\usepackage{tikz}

\newcommand{\mean}{\text{mean}}

\usepackage[numbers,sort&compress,sectionbib]{natbib}

\usepackage[nolist]{acronym}
\begin{acronym}[UML]
	\acro{KG}{Knowledge Graph}
    \acro{KGE}{Knowledge Graph Embedding}
    \acro{FOL}{First-Order Logic}
    \acro{AI}{Artificial Intelligence}
    \acro{EPFO}{Existential Positive First-order}
    \acro{DAG}{Directed Acyclic Graph}
    \acro{DNF}{Disjunctive Normal Form}
\end{acronym}  

\newcommand{\triple}[3]{(#1, #2, #3)}
\DeclareMathOperator*{\argmax}{arg\,max}

\usepackage{pifont}

\titlerunning{LitCQD: Multi-Hop Reasoning}

\begin{document}

\title{LitCQD: Multi-Hop Reasoning in Incomplete \texorpdfstring{\\}{} Knowledge Graphs with Numeric Literals}

\author{Caglar Demir \and
Michel Wiebesiek\and
Renzhong Lu\and
Axel-Cyrille Ngonga Ngomo\
Stefan Heindorf}
%
%
\institute{Paderborn University}

\maketitle

\begin{abstract}
Most real-world knowledge graphs, including Wikidata, DBpedia, and Yago are incomplete. Answering queries on such incomplete graphs is an important, but challenging problem. Recently, a number of approaches, including complex query decomposition (CQD), have been proposed to answer complex, multi-hop queries with conjunctions and disjunctions on such graphs. However, all state-of-the-art approaches only consider graphs consisting of entities and relations, neglecting literal values. In this paper, we propose LitCQD---an approach to answer complex, multi-hop queries where both the query and the knowledge graph can contain numeric literal values: LitCQD can answer queries having numerical answers or having entity answers satisfying numerical constraints. For example, it allows to query (1)~persons living in New York having a certain age, and (2)~the average age of persons living in New York. We evaluate LitCQD on query types with and without literal values. To evaluate LitCQD, we generate complex, multi-hop queries and their expected answers on a version of the FB15k-237 dataset that was extended by literal values.
\end{abstract}

\section{Introduction}

\acp{KG} such as Wikidata~\cite{vrandecic2014Wikidata}, DBpedia~\cite{auer2017DBpedia}, and YAGO~\cite{suchanek2007YAGO} have been of increasing interest in both academia and industry, e.g., for major question answering systems~\cite{diefenbach2017question, tahri2013dbpedia, adolphs2011YAGO-QA} and for intelligent assistants such as Amazon Alexa, Siri, and Google Now. 
Natural language questions on such \acp{KG} are typically answered by translating them into subsets of \ac{FOL} involving conjunctions ($\land$), disjunctions ($\lor$), and existential quantification ($\exists$) of multi-hop path expressions in the \acp{KG}. 
However, this approach to modeling queries has an important intrinsic flaw: 
Almost all real-world \acp{KG} are incomplete~\cite{nickel2016review, Farber2018Linked,dettmers2018convolutional}.
Traditional symbolic models, which rely on sub-graph matching, are 
unable to infer missing information on such incomplete \acp{KG}~\cite{hamilton2018embedding}. 
Hence, they often return empty answer sets to queries that can be answered by predicting missing information.
Hence, several approaches (e.g., GQE~\cite{hamilton2018embedding}, Query2Box~\cite{ren2020query2box}, and CQD~\cite{arakelyan2021complex}) have recently been proposed that can query incomplete \acp{KG} by performing neural reasoning over \acp{KGE}.
However, all the aforementioned models operate solely on \acp{KG} consisting of \emph{entities and relations} and none of them supports \acp{KG} with \emph{literal values} such as the age of a person, the height of a building, or the population of a city. Taking literal values into account, however, has been shown to improve predictive performance in many tasks~\cite{kristiadi2019incorporating,heindorf2022evolearner}.

In this paper, we remedy this drawback and propose LitCQD, a neural reasoning approach that can answer queries involving \emph{numerical literal values} over incomplete \acp{KG}.
LitCQD extends CQD by combining a \ac{KGE} model (e.g. ComplEx-N3~\cite{lacroix2018canonical}) that predicts missing entities/relations with a literal \ac{KGE} model (e.g. TransEA~\cite{Wu2018TransEA}) able to predict missing numerical literal values. 
Therewith, LitCQD can mitigate missing entities/relations as well as missing numerical values to answer various types of queries.
Moreover, we \emph{increase the expressiveness of queries} that can be answered on \acp{KG} with literal values by allowing queries (1) to contain filter restrictions involving literals and (2) to ask for predictions of numeric values (see~\Cref{example:first}). 
\begin{example}
The query ``\emph{Who ($P_?$) is married to somebody ($P$) younger than 25?}'' with a filter restriction ``younger than 25'' can be rewritten as
    $P_? . \exists P, C: \mathrm{hasAge}(P, C) \land \mathrm{lt}(C, 25) \land \mathrm{married}(P, P_?).$
    \label{example:first}
\end{example}
To answer this query, we predict the age of all persons $P$ in the knowledge graph and check whether the condition ``less than 25'' is fulfilled. Then, all persons $P_?$ married to persons $P$ are returned.

A particular challenge was to develop an efficient continuous counterpart to discrete, Boolean filter expressions such as ``less than 25'' that works on incomplete knowledge graphs. To this end, we introduce continuous attribute filter functions (Section~\ref{subsec:answering_entity_quers}, Equations~\ref{eq:attribute_filter_equal}--\ref{eq:attribute_filter_gt}) and improve them by introducing attribute existence checks (Equations~\ref{eq:preliminary_attribute_filter_predictor}--\ref{eq:final_attribute_filter_predictor}). Another challenge was predicting attribute values for a subset of entities specified by a query on an incomplete knowledge graph. We predict attribute values by means of a beam search over entities obtained via attribute filter functions (Section~\ref{subsec:literal_value_queries}). We experimented with several variants, but due to space constraints, we focus on the best-performing one in this paper and mention alternative variants only briefly.

In our experiments, we use a similar setup to \citet{hamilton2018embedding,garcia2018kblrn,arakelyan2021complex} and use the FB15k-237 dataset augmented with literals~\cite{garcia2018kblrn}. However, as previous work did not contain queries with literal values, we generate such queries and their expected answers. Our experiments suggest that LitCQD can effectively answer various types of queries involving literal values, which was not possible before (Tables~\ref{tab:eval_query_results_attr_filtering}, \ref{tab:eval_query_results_attr_pred}). 
Moreover, our results show that including literal values during the training process improves the query answering performance even on standard queries in our benchmark (Table~\ref{tab:eval_queries_without_literals}). Our contributions can be summarized as follows:
\begin{itemize}
    \item \emph{Filter restrictions with literals:} We propose an approach that can answer multi-hop queries where numeric literals are used to filter valid answers (e.g., ``return entities whose age is less than 25'')
    \item \emph{Prediction of literal values:} We propose an approach that can \emph{predict the numeric values} of literals (e.g., ``return mean age of married people').
    \item \emph{Benchmark construction:} We generate multi-hop queries \emph{with numeric literals} and their expected answers
    \item \emph{Embeddings with literals:} We show that using knowledge graph embeddings that support literal values even yields better results for traditional queries without literal values
\end{itemize}




\section{Background and Preliminaries}\label{sec:background}
In this section, we give a brief introduction to knowledge graphs without literals and queries on knowledge graphs without literals, before introducing our approach for knowledge graphs with literals in \Cref{sec:methodology}.

\subsection{Knowledge Graph without Literals}
\label{subsec:knowledge_graph}
A knowledge graph (\ac{KG}) without literals is defined as $\mathcal{G} = \{ \triple{h}{r}{t} \} \subseteq \mathcal{E} \times \mathcal{R} \times \mathcal{E}$, where $h,t \in \mathcal{E}$ denote entities and
$r \in \mathcal{R}$ denotes a relation~\cite{hamilton2018embedding,ren2020query2box}.
$\mathcal{G}$ can be regarded as a \ac{FOL} knowledge base, where a relation $r \in \mathcal{R}$ corresponds to a binary function $\hat{r}:\mathcal{E} \times \mathcal{E} \rightarrow \{1, 0\}$ and a triple $(h,r,t)$ corresponds to an atomic formula $\alpha = \hat{r}(h, t)$~\cite{arakelyan2021complex}. When it is clear from the context that $\hat{r}$ denotes a binary function, we may simply write $r$ as in the following definitions.

\subsection{Multihop Queries without Literals}

\paragraph{Conjunctive Queries.}
A conjunctive graph query~\cite{hamilton2018embedding,ren2020beta,arakelyan2021complex, ren2020query2box} $q \in \mathcal{Q}(\mathcal{G})$ over $\mathcal{G}$ is defined as
\begin{equation}
    q = E_?\:.\:\exists E_1, \ldots, E_{m} : \alpha_1\land \alpha_2 \land \ldots \land \alpha_n,
    \label{eq:cgq}
\end{equation}
where
\begin{itemize}
    \item $\textrm{$\alpha_i = r(e, E)$}$, with $E \in \{E_?,E_1, \dots, E_{m}\}$, $r \in \mathcal{R}$, $e \in \mathcal{E}$ or
    \item $\textrm{$\alpha_i = r(E, E^{\prime})$}$, with $E, E' \in \{E_?, E_1, \dots, E_{m}\}$, $E\neq E'$, $r \in \mathcal{R}$.
\end{itemize}
In the query, the target variable $E_?$ and the existentially quantified variables $E_1, \dots, E_m$ are bound to subsets of \emph{entities} $\mathcal{E}$. The entities bound to $E_?$ represent the answer nodes of the query.
The conjunction $\alpha_1\land \alpha_2 \land \ldots \land \alpha_n$ consists of $n$ atoms defined over relations $r\in \mathcal{R}$, anchor entities $e \in \mathcal{E}$ and variables $E, E'\in \{ E_?, E_{1}, \ldots, E_{m} \}$.


%
\begin{example}
To give a concrete example, the natural language question ``\emph{Which ($D_?$) drugs are to interact with ($P$) proteins associated with the diseases $e_1$ and $e_2$?}'' can be represented as the conjunctive graph query
\begin{equation}
\label{eq:conjunctive_query}
    q=D_?.\exists P : \mathit{assoc}(e_1, P) \land \mathit{assoc}(e_2,P) \land \mathit{interacts}(P,D_?),
\end{equation}
where 
$D_?,P$ are bound to subsets of entities $\mathcal{E}$,
$e_1, e_2 \in \mathcal{E}$ are anchor entities and $\mathit{interacts}, \mathit{assoc} \in \mathcal{R}$ are relations.
\end{example}

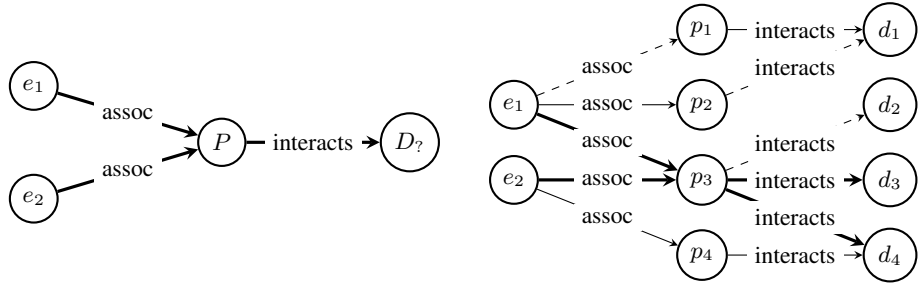
\begin{figure}[tb]
    \begin{minipage}{0.47\textwidth}
    \centering
    \begin{tikzpicture}
        \begin{scope}[every node/.style={circle,thick,draw}]
            \node (1) at (0,1.5) {$e_1$};
            \node (2) at (0,0) {$e_2$};
            \node (3) at (2.5,0.75) {$P$};
            \node (4) at (5,0.75) {$D_?$};
        \end{scope}
        \begin{scope}[>={stealth[black]},
            every node/.style={fill=white},
            every edge/.style={draw=black,very thick}]
            \path [->] (1) edge node {assoc} (3);
            \path [->] (2) edge node {assoc} (3);
            \path [->] (3) edge node {interacts} (4);
        \end{scope}
    \end{tikzpicture}
    \end{minipage}\quad\quad%
    \begin{minipage}{0.47\textwidth}
    \centering
    \begin{tikzpicture}
        \begin{scope}[every node/.style={circle,thick,draw}]
            \node (e1) at (0,1) {$e_1$};
            \node (e2) at (0,0) {$e_2$};
            
            \node (p1) at (2.5,2) {$p_1$};
            \node (p2) at (2.5,1) {$p_2$};
            \node (p3) at (2.5,0) {$p_3$};
            \node (p4) at (2.5,-1) {$p_4$};

            \node (d1) at (5,2) {$d_1$};
            \node (d2) at (5,1) {$d_2$};
            \node (d3) at (5,0) {$d_3$};
            \node (d4) at (5,-1) {$d_4$};
        \end{scope}
        \begin{scope}[>={stealth[black]},
            every node/.style={fill=white},
            every edge/.style={draw=black}]
            \path [->] (e1) edge node {assoc} (p2);
            \path [->] (e2) edge node {assoc} (p4);
            \path [->] (p1) edge node {interacts} (d1);
            \path [->] (p4) edge node {interacts} (d4);
        \end{scope}
        \begin{scope}[>={stealth[black]},
            every node/.style={fill=white},
            every edge/.style={draw=black,dashed}]
            \path [->] (e1) edge node {assoc} (p1);
            \path [->] (p2) edge node {interacts} (d1);
            \path [->] (p3) edge node {interacts} (d2);
        \end{scope}
        \begin{scope}[>={stealth[black]},
            every node/.style={fill=white},
            every edge/.style={draw=black,very thick}]
            \path [->] (e1) edge node {assoc} (p3);
            \path [->] (e2) edge node {assoc} (p3);
            \path [->] (p3) edge node {interacts} (d3);
            \path [->] (p3) edge node {interacts} (d4);
        \end{scope}
    \end{tikzpicture}
    \end{minipage}
    \caption{Example query without literals (see \Cref{eq:conjunctive_query}). Dependency graph of query (left) and symbolic query answering on an incomplete graph (right). Solid bold lines represent paths leading to answer entities. Dashed lines represent missing triples.}
    \label{fig:example-query-without-literals}
\end{figure}

The dependency graph of a query $q \in Q(\mathcal{G})$ is defined over its query edges $\alpha_1,\allowbreak{} \alpha_2,\allowbreak{} \ldots,\allowbreak{} \alpha_n$ with nodes being either anchor entities or variables~\cite{hamilton2018embedding}. Following \citet{hamilton2018embedding, arakelyan2021complex}, we focus on queries whose dependency graph forms a \ac{DAG} with anchor entities being source nodes and the target variable being the unique sink node (such queries are called \emph{valid} queries in previous work~\cite{hamilton2018embedding,arakelyan2021complex}).
For example, \Cref{fig:example-query-with-literals} (left) represents the dependency graph of the query in~\Cref{eq:conjunctive_query}. Note that for the sake of brevity, we use the term of entity in a $\mathcal{G}$ interchangeably with a node in a dependency graph.



%
The dependency graph of a query encodes the \emph{computation graph} to obtain the answer set $\llbracket q \rrbracket$ via
\emph{projection}~$\mathcal{P}$ and 
\emph{intersection}~$\mathcal{I}$ operators~\cite{ren2020query2box}.
Starting from a set of anchor nodes (e.g. $e_1, e_2$),
$\llbracket q \rrbracket$ is derived by iteratively applying
$\mathcal{P}$ and/or $\mathcal{I}$ until the unique sink target node (e.g. $D_?$) is reached.
Given a set of entities $S \subseteq \mathcal{E}$ and a relation $r \in \mathcal{R}$, the projection operator is defined as 
$\mathcal{P}(S,r):=\cup_{e \in S} \;
\{x \in \mathcal{E}: \ \hat{r}(e, x) = 1\}$ where the binary function $\hat{r}: \mathcal{E} \times \mathcal{E} \rightarrow \{1, 0\}$ indicates whether the triple $(e, r, x)$ exists in $\mathcal{G}$.%
\footnote{When computing the ground truth answer on the complete graph, we check whether $(e,r,x) \in \mathcal{G}$ (see details on query generation below and in \citet{hamilton2018embedding}). When performing neural reasoning, $\hat{r}$ is approximated with a link predictor yielding a score between 0 and 1.}
Given a set of entity sets $\{S_1, S_2, \dots, S_n \}, S_i \subseteq \mathcal{E}$, the intersection operator $\mathcal{I}$ is defined as
$\mathcal{I}(\{S_1, S_2, \dots, S_n \}):= \cap_{i=1}^{n} S_i$.
Therefore, the conjunctive query defined in ~\Cref{eq:conjunctive_query} can be answered via the computation %
\begin{equation}
\mathcal{P}\Big(\mathcal{I}\big(\big\{\mathcal{P}(\{e_1\}, \mathit{assoc}), \mathcal{P}(\{e_2\}, \mathit{assoc})\big\}\big), \mathit{interacts}\Big).
\label{eq:cqd_answer_dependency}
\end{equation}
In the example of \Cref{fig:example-query-with-literals} (right), a traditional, symbolic approach yields the answer set $\llbracket q \rrbracket= \{d_3,d_4\}$ although the complete answer set taking missing triples into account would be $\llbracket q \rrbracket= \{d_2, d_3,d_4\}$. The result is obtained as follows: Starting at the anchor entities $e_1$ and $e_2$, the entity $p_3$ is the only entity for which both $\mathit{assoc}(e_1, p_3)$ and $\mathit{assoc}(e_2, p_3)$ hold. Moving on from $p_3$, a traditional, symbolic approach can only reach the entities $d_3, d_4$ via the ``interacts'' relation, but not the entity $d_2$ because the edge $(p_3, \text{interacts}, d_2)$ is missing. Note that $d_1$ is not part of the answer set because both $p_1$ and $p_2$ are only associated with $e_1$.

\paragraph{\ac{EPFO} Queries.}
An \ac{EPFO} query $q$ in its \ac{DNF} is a disjunction of conjunctive queries~\cite{ren2020query2box,arakelyan2021complex}:
\begin{equation}
    q = E_?\:.\:\exists E_1, \dots, E_{m} : (\alpha_1^1\land \dots \land \alpha_{n_1}^1)\lor \dots \lor (\alpha_{n_1}^d\land \dots \land \alpha_{n_d}^d), 
    \label{eq:epfo_query}
\end{equation}
where $\alpha_i^j$ are defined as above.
Its dependency graph is a \ac{DAG} having three types of directed edges: \emph{projection}, \emph{intersection}, and \emph{union}; the union $\mathcal{U}$ of entity sets
$S_1, S_2, \dots,\allowbreak{} S_n \subseteq \mathcal{E}$ is
$\mathcal{U}(\{S_1, S_2, \dots, S_n \}):= \cup_{i = 1}^n S_i.$

\section{Related Work}
\label{sec:related-work}
In this section, we overview the state of the art with regards to knowledge graph embeddings and neural query answering on incomplete knowledge graphs.

\subsection{Knowledge Graph Embeddings and Literals}
\label{subsec:kge_and_literals}
In the last decade, a plethora of knowledge graph embedding (\ac{KGE}) models have been successfully applied to tackle various tasks, including link prediction, relation prediction, community detection, fact checking, and class expression learning~\cite{nickel2016review,trouillon2016complex,silva2021using,kouagou2022learning}.
\ac{KGE} research has mainly focused on learning embeddings for entities and relations tailored towards predicting missing entity/relation given a triple, i.e., tackling single-hop queries~\cite{nickel2016review,trouillon2016complex,yang2015embedding,dettmers2018convolutional,balavzevic2019tucker,sun2019rotate,zhang2019quaternion,demir2021hyperconvolutional,demir2021convolutional}. 
Despite their effectiveness in tackling single-hop queries, \ac{KGE} models cannot be directly applied to answer multi-hop queries.
This is due to the fact that multi-hop query answering over \acp{KG} is a strict generalization of knowledge graph completion (i.e., single-hop query answering)~\cite{ren2022smore}.
Moreover, most \ac{KGE} do not incorporate literals (e.g. numeric attributes) in \acp{KG}. 
Consequently, embeddings for entities and relations are learned without incorporating knowledge encoded with literals (e.g., age of a person, height of a person or date of birth).
To alleviate this limitation, there has been a growing interest in designing \ac{KGE} model incorporating literals in recent years.
For instance, Wu et al.~\cite{wu2018knowledge} propose TransEA by extending the translation loss used in TransE~\cite{bordes2013translating} by adding the attribute loss as a weighted regularization term. 
Garcia-Duran and Niepert~\cite{garcia2018kblrn} propose KBLRN that is based on relation features, numerical literals, and a \ac{KGE} model.
A predicted score of a triple is composed of relation feature values, predicted scores via a \ac{KGE} model, and a numerical literal feature.
A relation feature is a logical 2-hop formula (e.g. $\exists x : \mathit{bornIn}(a,x) \land 
\mathit{capitalOf}(x,b)$) generated by AMIE+~\cite{galarraga2015fast} that acts as a binary classifier and assigns 1 if there is a path from an entity $a$ to $b$, otherwise 0.
A literal feature is constructed by taking the difference between a numeric value of subject and object entities for a given relation.
\citet{garcia2018kblrn} show that the mean differences of birth years is 0.4 on Freebase between entities occurring with \texttt{/people/marriage/spouse}, whereas it is 32.4 for the relation \texttt{/person/children}.
\citet{kristiadi2019incorporating} propose LiteralE that applies a non-linear parameterized function to merge entity embeddings with numerical literals.
By this, LiteralE is computationally less demanding than KBLRN as it does not require any rule generation and is more expressive than TransE as TransE integrates the impact of literals linearly.
Learning a parameterized function to enrich entity embeddings with their numerical literal information available in \acp{KG} improves the link prediction performance across benchmark datasets.

\subsection{Neural Query Answering on Incomplete Knowledge Graphs}
\label{subsec:qa_kg}

In recent years, significant progress has been made on querying incomplete \acp{KG}.
\citet{hamilton2018embedding} laid the foundations for multi-hop reasoning with graph query embeddings (GQE). Given a conjunctive query (e.g. \Cref{eq:conjunctive_query}), 
they learn continuous vector representations for queries, entities, and relations. Queries on incomplete knowledge graphs are answered by performing projection $\mathcal{P}$ and intersection $\mathcal{I}$ operations in the embedding vector space. \citet{ren2020query2box} show that GQE cannot answer \ac{EPFO} queries (see~\Cref{eq:epfo_query}) since GQE does not model the union operator $\mathcal{U}$.
To answer \ac{EPFO} queries in \acp{DNF},~\citet{ren2020query2box} propose Query2Box that represents an \ac{EPFO} query with a set of box embeddings, where a one box embedding is constructed per conjunctive subquery. 
A query is answered by returning the entities whose minimal distance to one of the box embeddings is smallest.

All the aforementioned models learn query embeddings and answer queries via nearest neighbor search in the embedding space.
However, learning embeddings for complex, multi-hop queries involving conjunctions and disjunctions can be computationally demanding. 
Towards this end, \citet{arakelyan2021complex} propose complex query decomposition (CQD). 
They answer \ac{EPFO} queries by decomposing them into single-hop subqueries and aggregate the scores of a pre-trained single-hop link predictor (e.g. ComplEx-N3). 
Scores are aggregated using a t-norm and t-conorm---continuous generalizations of the logical conjunction and disjunction~\cite{arakelyan2021complex,klement2004triangular}.
Their experiments suggest that CQD outperforms GQE and Query2Box; it generalizes well to complex query structures while requiring orders of magnitude less training data.
Zhu et al.~\cite{zhu2022neural} highlight that CQD is the only interpretable model among the aforementioned models as it produces intermediate results.
In this work, we extend CQD to answer multi-hop queries involving literals.


\section{LitCQD: Multi-hop Reasoning with Literals}
\label{sec:methodology}

A knowledge graph with numeric literals (i.e. with scalar values), can be defined as $\mathcal{G}_A = \{ \triple{h}{r}{t} \} \subset (\mathcal{E} \times \mathcal{R} \times \mathcal{E}) \cup (\mathcal{E} \times \mathcal{A} \times \mathbb{R})$, where
$\mathcal{R} \cap \mathcal{A}=\emptyset $ and $\mathcal{A}$ and $\mathbb{R}$ denote numeric attributes and real numbers, respectively~\cite{kristiadi2019incorporating}. The binary function $\hat{a}:\mathcal{E} \times \mathbb{R} \mapsto \{1, 0\}$ indicates whether an entity has attribute $a \in \mathcal{A}$ and we might just write $a$ instead of $\hat{a}$ when this is clear from context.
We categorize \ac{EPFO} queries $q \in \mathcal{Q}(\mathcal{G}_A)$ involving literals depending on the type of their answer sets $\llbracket q \rrbracket$: In~\Cref{subsec:answering_entity_quers}, we define queries with entities as answer set $\llbracket q \rrbracket \subseteq \mathcal{E}$;
in~\Cref{subsec:literal_value_queries}, we define queries with a literal value as answer $\llbracket q \rrbracket \in \mathbb{R}$.

\subsection{Multihop Queries with Literals and Entity Answers}
\label{subsec:answering_entity_quers}

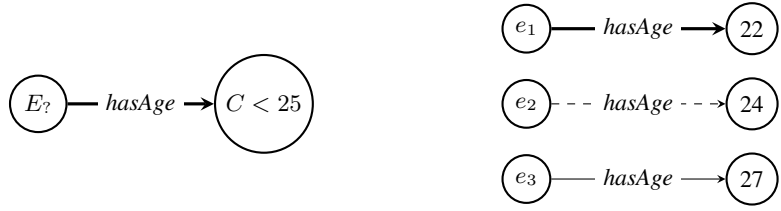
\begin{figure}[tb]
    \begin{minipage}{0.47\textwidth}
    \centering
    \begin{tikzpicture}
        \begin{scope}[every node/.style={circle,thick,draw}]
            \node (1) at (0,0) {$E_?$};
            \node (2) at (3,0.0) {$C<25$};
        \end{scope}
        \begin{scope}[>={stealth[black]},
            every node/.style={fill=white},
            every edge/.style={draw=black,very thick}]
            \path [->] (1) edge node {\emph{hasAge}} (2);
        \end{scope}
    \end{tikzpicture}
    \end{minipage}\quad\quad%
    \begin{minipage}{0.47\textwidth}
    \centering
    \begin{tikzpicture}
        \begin{scope}[every node/.style={circle,thick,draw}]
            \node (e1) at (0,1.0) {$e_1$};
            \node (e2) at (0,0) {$e_2$};
            \node (e3) at (0,-1.0) {$e_3$};
            \node (age22) at (3,1) {22};
            \node (age24) at (3,0) {24};
            \node (age27) at (3, -1) {27};
        \end{scope}
        \begin{scope}[>={stealth[black]},
            every node/.style={fill=white},
            every edge/.style={draw=black}]
            \path [->] (e3) edge node {\emph{hasAge}} (age27);
        \end{scope}
        \begin{scope}[>={stealth[black]},
            every node/.style={fill=white},
            every edge/.style={draw=black,dashed}]
            \path [->] (e2) edge node {\emph{hasAge}} (age24);
        \end{scope}
        \begin{scope}[>={stealth[black]},
        every node/.style={fill=white},every edge/.style={draw=black,very thick}]            
            \path [->] (e1) edge node {\emph{hasAge}} (age22);
        \end{scope}
    \end{tikzpicture}
    \end{minipage}
    \caption{Example query with literals and entity answer (see \Cref{eq:conjunctive_query_with_age}). On the left, the query's dependency graph is shown and on the right, symbolic query answering on an incomplete graph with literal values. Bold lines represent paths leading to answer entities, dashed lines represent missing triples, solid existing triples.}
    \label{fig:example-query-with-literals}
\end{figure}

An \ac{EPFO} query $q$ on a knowledge graph with numeric literals ($\mathcal{G}_\mathcal{A}$) can be defined as
\begin{equation}
    q = E_?\:.\:\exists E_1, \dots,E_{m} :   (\alpha_1^1\land \dots \land \alpha_{n_1}^1)\lor \dots \lor
    (\alpha_1^d\land \dots \land \alpha_{n_d}^d),
\label{eq:epfo_with_attributes}
\end{equation}
where
\begin{itemize}
    \item $\alpha_i^j=r(e, E)$, with $E \in \{E_?,E_1, \dots, E_{m}\}$, $r \in \mathcal{R}$, $e \in \mathcal{E}$ or
    \item $\alpha_i^j = r(E, E')$, with $E, E' \in \{E_?, E_1, \dots, E_{m}\}$, $E\neq E'$, $r \in \mathcal{R}$~or
    \item $\alpha_i ^j = a(E, C) \land \textit{af}(C, c)$, with $E \in \{E_?,E_1, \dots, E_{m}\}$, $C \in \{C_1, \dots, C_l\}$
    $a \in \mathcal{A}$, $\textit{af} \in \{\mathrm{lt}, \mathrm{gt}, \mathrm{eq}\}$, $c \in \mathbb{R}$.
\end{itemize}
In the query, the target variable $E_?$ and the variables $E_1, \dots, E_m$ are bound to subsets of \emph{entities} $\mathcal{E}$ and the variables $C_1, \dots, C_l$ are bound to numeric values from $\mathbb{R}$. The binary function $r: \mathcal{E} \times \mathcal{E} \mapsto \{1,0\}$ denotes whether a relation exists between the two entities, $a:\mathcal{E} \times \mathbb{R} \mapsto \{1,0\}$ denotes whether an attribution relation exists, and $\textit{af}: \mathbb{R} \times \mathbb{R} \mapsto \{1,0\}$ is one of the attribute filter conditions lt (\emph{less-than}), gt (\emph{greater-than}), or eq (\emph{equal-to}). For example, $\mathit{lt}(20, 25)$ returns 1 because $20 \leq 25$. To approximately answer queries defined with~\Cref{eq:epfo_with_attributes} and assuming an incomplete knowledge graph, we propose the following optimization problem:
\begin{equation} 
\argmax_{E_?, E_1, \ldots, E_m}  \left( \alpha^{1}_{1} 
\ \top \ \dots \ \top \ \alpha^{1}_{n_1} \right) \ \bot \ \dots \ \bot \ \left( \alpha^{d}_{1} \ \top \ \dots \ \top \ \alpha^{d}_{n_d} \right)
\label{eq:optimization_problem_entity_answers}
\end{equation}
where 
\begin{itemize}
    \item $\alpha^{j}_{i}=\phi_r(e,E)$, with $E \in \{ E_?, E_{1}, \ldots, E_{m} \}$, $r \in \mathcal{R}$, $e \in \mathcal{E}$ or
    \item $\alpha^{j}_{i}=\phi_r(E,E')$, with $E,E'\in \{ E_?, E_{1}, \ldots, E_{m} \}$, $E\neq E'$, $r \in \mathcal{R}$ or
    \item $\alpha^{j}_{i}=\phi_{\textit{af}, a}(\phi_a(E),c)$, with $E \in \{ E_?, E_{1}, \ldots, E_{m} \}, c\in \mathbb{R}$,
\end{itemize}
and $\phi_r: \mathcal{E} \times \mathcal{E} \mapsto [0,1]$ is a link predictor that predicts a \emph{likelihood} of a link between two entities via a relation $r$.
$\phi_a: \mathcal{E} \mapsto \mathbb{R}$ is an attribute
predictor that predicts a \emph{value} of an attribute $a$ given an entity.
An attribute filter predictor $\phi_\textit{af, a}: \mathbb{R} \times \mathbb{R} \mapsto [0,1]$ predicts a \emph{likelihood} that the filter condition is met given the predicted attribute value $\hat{c}:=\phi_a(\cdot)$ and the constant value $c \in \mathbb{R}$ specified in the query.
All three predictors are derived from a \ac{KGE} model as described below.
A t-norm $\top: [0,1] \times [0,1] \mapsto [0,1]$ is considered as a continuous generalization of the logical conjunction~\cite{arakelyan2021complex,klement2004triangular}.
Given a t-norm $\top$, the complementary t-conorm can be defined as $\bot(a,b)= 1- \top(1-a,1-b)$~\cite{arakelyan2021complex}.
Numerically, the \emph{Gödel t-norm} $\top_{\text{min}}(x, y) = \min\{ x, y \}$, the \emph{product t-norm} $\top_{\text{prod}}(x, y) = x \cdot y$, or the \emph{Łukasiewicz t-norm} $\top_{\text{Luk}}(x, y) = \max \{ 0, x + y - 1 \}$ can be used to aggregate predicted likelihoods to obtain a query score~\cite{arakelyan2021complex}.
With this formulation, various questions involving numerical values can be asked on incomplete~$\mathcal{G}_\mathcal{A}$.
For example, the question ``\emph{Which entities are younger than 25?}'' can be represented as
\begin{equation}
\label{eq:conjunctive_query_with_age}
q=  E_?\:.\:\exists C: \mathit{hasAge}(E_?, C)\land \emph{lt}(C,25). 
\end{equation}
The dependency graph of this query $q$ is visualized in~\Cref{fig:example-query-with-literals} (left). Let $S_?$ be the entities bound to variable $E_?$. Then the projection of $S_?$ with $\emph{hasAge}$ is performed by an attribute prediction model $\phi_\emph{hasAge}(S_?) \in \mathbb{R}^{|E|}$ that predicts the value of the attribute~$a$ for each entity in $e \in E$.
Then the answer set is obtained by filtering entities via $\phi_\mathrm{lt}$. A subgraph in $\mathcal{G}_\mathcal{A}$ satisfying this query is visualized in~\Cref{fig:example-query-with-literals} (right). While a symbolic approach would only yield the answer set $\llbracket q \rrbracket = \{e_1\}$, our approach involving link predictors can identify the full answer set $\llbracket q \rrbracket = \{e_1, e_2\}$.

We solve the optimization problem in \Cref{eq:optimization_problem_entity_answers} approximately with a variant of beam search by greedily searching for sets of entities $S_?, S_1, \ldots S_m$ substituting the variables ${E_?, E_1, \ldots, E_m}$ in a fashion akin to CQD~\cite{arakelyan2021complex}.
In the example in \Cref{eq:conjunctive_query_with_age}, given the $\emph{hasAge}$ attribute, attribute values $\hat{c} = \phi_\emph{hasAge}(e) \in \mathbb{R}$ are predicted for all entities $e \in \mathcal{E}$.%
\footnote{This operation can be done \emph{in a single step} on a GPU by using the entity embedding matrix~\cite{arakelyan2021complex}.}
Next, likelihoods of fulfilling the filter condition ``less than 25'' can be inferred via $\phi_\mathrm{lt}(\hat{c},25)$.
Finally, all entities are sorted by their query scores in descending order and the top $k$ entities are considered to be answers of $q$.



It is important to note that LitCQD like CQD not only computes the final answer but also intermediate steps leading to this answer. In this sense, LitCQD can be considered an interpretable model.

%
\paragraph{Joint Training of Link and Attribute Predictors.}

Following \citet{arakelyan2021complex}, we use ComplEx-N3~\cite{lacroix2018canonical} as entity predictor $\phi_r(\cdot, \cdot)$. 
As attribute predictor $\phi_a(\cdot)$, we employ TransEA~\cite{Wu2018TransEA}. We jointly train the \ac{KGE} models underlying both models.

The link predictor ComplEx-N3 has previously been found to work well for multi-hop query answering~\cite{arakelyan2021complex} and to perform better than DistMult~\cite{yang2015embedding, arakelyan2021complex}. In a pilot study, we also experimented with the attribute predictor MTKGNN~\cite{Tay2017MTKGNN}. Overall, it achieved similar performance to TransEA, but we decided to move forward with TransEA, because it slightly outperformed MTKGNN in terms of MRR and required less parameters. KBLN~\cite{garcia2018kblrn} and LiteralE~\cite{kristiadi2019incorporating} can only be used to compute knowledge graph embeddings based on literal information, but they do not allow to predict the value of attributes which is required in our framework. 




\paragraph{Attribute Filter Function without Existence Check.}
The attribute filter function returns a score indicating the likelihood that the filter condition is met. First, we define a preliminary version $\phi'_{\textit{af}, a}$ of the function, which does not check whether the attribute relation $a$ actually exist for an entity. The function is defined case by case as described in the following. For the \emph{equal-to} condition, i.e., for $\textit{af}=\mathrm{eq}$, we define it as
\begin{equation}
\phi'_{\mathrm{eq}, a}(\hat{c},c) := \frac{1}{\mathrm{exp}({| \hat{c}-c | / \sigma_a})},
\label{eq:attribute_filter_equal}
\end{equation}
where $\hat{c}=\phi_a(e), e \in \mathcal{E}$, $c \in \mathbb{R}$ is a numeric literal (e.g. 25 in~\Cref{fig:example-query-with-literals}, left) and $\sigma_a$ denotes the standard deviation of $\mathcal{C}_a$ where $\mathcal{C}_a:=\{c \in \mathbb{R}|\hat{a}(e, c)=1, e \in \mathcal{E}\}$ are all literal values found on $\mathcal{G}_\mathcal{A}$ given an attribute $a$.
With $\phi_{\mathrm{eq},a}(\hat{c},c)$, we map the difference between the predicted attribute value $\hat{c}$ and the constant value $\hat{c}$ specified in the query into the unit interval $[0,1]$.
As the difference $\mid \hat{c}-c \mid$ approaches $0$, $\phi_{\mathrm{eq},a}(\hat{c},c)$ approaches $1$.
The division by the standard deviation $\sigma$ normalizes the difference $\mid \hat{c}-c \mid$.
For the attribute filter function with $\emph{less-than}$ ($\textit{af}=\mathrm{lt}$), we define
\begin{equation}
\phi'_\mathrm{lt}(\hat{c},c) :=\frac{1}{ 1+\mathrm{exp}((\hat{c}-c)/\sigma_a)}.
\label{eq:attribute_filter_gt}
\end{equation}
As $\hat{c}-c \to -\infty$, $\phi_\mathrm{lt}(\hat{c},c) \to 1$.
Following~\Cref{eq:attribute_filter_gt},
the attribute filter function with $\emph{greater-than}$ is defined as
\begin{equation}
\phi'_\mathrm{gt}(\hat{c},c) := 1- \phi_\mathrm{lt}(\hat{c},c).
\label{eq:attribute_filter_lt}
\end{equation}
We also experimented with a version where the standard deviation $\sigma_\alpha$ was not computed per attribute but for all literal values in the knowledge graph.


\paragraph{Attribute Filter Function with Existence Check.}
The preliminary attribute filter function $\phi'_{\textit{af, a}}$ assumes that the attribute relation $a$ exists for each entity in the knowledge base which is clearly not the case. Hence, we employ a model $\phi_{\mathrm{exists},a}(e)$ that scores the likelihood that the attribute relation $a$ exists for entity $e$. Then the final attribute filter function $\phi_{\textit{af, a}}$ is obtained by combining the attribute existence predictor $\phi_{\mathrm{exist}, a}(e)$ with the preliminary filter predictor $\phi'_{\textit{af, a}}$:
\begin{equation}
    \phi_{\textit{af}, a}(\hat{c}, c) := \phi_{\mathrm{exists},a}(e) \cdot \phi'_{\textit{af, a}}(\hat{c}, c)
    \label{eq:preliminary_attribute_filter_predictor}
\end{equation}
Technically, the attribute existence predictor is realized by adding a dummy entity~$e_\mathrm{exists}$ to the knowledge base along with dummy edges $r_a(e, e_\mathrm{exists})$ if entity $e$ has an attribute relation $a$. Then, the existence of an attribute is predicted with the link predictor as
\begin{equation}
    \phi_{\mathrm{exists}, a}(e):= \phi_{r_a}(e, e_\mathrm{exists})
    \label{eq:final_attribute_filter_predictor}
\end{equation}
Note that the dummy entity and the dummy relations are only added to the train set but not the validation or test set.

\subsection{Multihop Queries with Literals and Literal Answers}
\label{subsec:literal_value_queries}

Here, we define an \ac{EPFO} query $q$ on an incomplete $\mathcal{G}_\mathcal{A}$, whose answer $\llbracket q \rrbracket \in \mathbb{R}$ is a real number (instead of a subset of entities) as follows
\begin{equation}
    q = \psi(C_?)\:.\:\exists E_?, E_1, \dots,E_{m} : (\alpha_1^1\land \dots \land \alpha_{n_1}^1)\lor \dots \lor (\alpha_1^d\land \dots \land \alpha_{n_d}^d),
\label{eq:epfo_with_attributes_and_real_num_naser}
\end{equation}
where $\psi:2^\mathbb{R} \mapsto \mathbb{R}$ is a permutation-invariant aggregation function and
\begin{itemize}
    \item $\alpha_i^j=r(e, E)$, with $E \in \{E_?, E_1, \dots, E_{m}\}$, $r \in \mathcal{R}$, $e \in \mathcal{E}$ or
    \item $\alpha_i^j = r(E, E')$, with $E, E' \in \{E_?, E_1, \dots, E_{m}\}$, $E\neq E'$, $r \in \mathcal{R}$~or
    \item $\alpha_i ^j = a(E, C) \land \textit{af}(C, c)$, with $E \in \{E_?, E_1, \dots, E_{m}\}$, $C \in \{C_?, C_1, \dots, C_l\}$
    $a \in \mathcal{A}$, $\textit{af} \in \{\mathrm{lt}, \mathrm{gt}, \mathrm{eq}\}$, $c \in \mathbb{R}$.
\end{itemize}
Variable bindings $S_?, S_1, \ldots, S_m$ for $E_?, E_1, \dots, E_m$ are obtained via the same optimization problem as in \Cref{subsec:answering_entity_quers}. Then the set of values $C_?$ can be computed by applying the attribute value predictor $\phi_a$ on the entities in $S_?$.

\begin{figure}[tb]
    \centering
    \begin{minipage}{0.47\textwidth}
    \centering
        \begin{tikzpicture}
        \begin{scope}[every node/.style={circle,thick,draw}]
            \node (1) at (2.45,0) {$E_?$};
            \node (2) at (4.9,0) {$C_?$};
            \node (3) at (0,0) {TA};
        \end{scope}
        \begin{scope}[>={stealth[black]},
            every node/.style={fill=white},
            every edge/.style={draw=black,very thick}]
            \path [->] (1) edge node {\emph{hasAge}} (2);
            \path [->] (1) edge node {\emph{winner}} (3);
        \end{scope}
    \end{tikzpicture}
    \end{minipage}\quad\quad%
    \begin{minipage}{0.47\textwidth}
    \centering
        \begin{tikzpicture}
        \begin{scope}[every node/.style={circle,thick,draw}]
            \node (e1) at (2.5,1.0) {$e_1$};
            \node (e2) at (2.5,0) {$e_2$};
            \node (ta) at (0,.5) {TA};
            \node (e1_age) at (5,1) {$22$};
            \node (e2_age) at (5,0) {$24$};
        \end{scope}
        \begin{scope}[>={stealth[black]},
            every node/.style={fill=white},
            every edge/.style={draw=gray}]
        \end{scope}
        \begin{scope}[>={stealth[black]},
            every node/.style={fill=white},
            every edge/.style={draw=black,dashed,very thick}]
            \path [->] (e2) edge node {\emph{winner}} (ta);
            
        \end{scope}
        \begin{scope}[>={stealth[black]},
            every node/.style={fill=white},every edge/.style={draw=black,very thick}]            
            \path [->] (e1) edge node {winner} (ta);
            \path [->] (e2) edge node {winner} (ta);
            \path [->] (e1) edge node {hasAge} (e1_age);
        \end{scope}
        \begin{scope}[>={stealth[black]},
            every node/.style={fill=white},
            every edge/.style={draw=black,dashed}]
            \path [->] (e2) edge node {hasAge} (e2_age);
        \end{scope}
    \end{tikzpicture}
    \end{minipage}
    \caption{Example of a query predicting attribute values (see \Cref{eq:turing_award_winners_avg_ge}). On the left, the dependency graph of the query is shown, on the right a subgraph to answer $q$. 
    Dashed lines represent missing information. 
    Bold lines represent paths leading to the symbolic answer $\llbracket q \rrbracket = 22$.}
    \label{fig:example-query-with-value-prediction}
\end{figure}
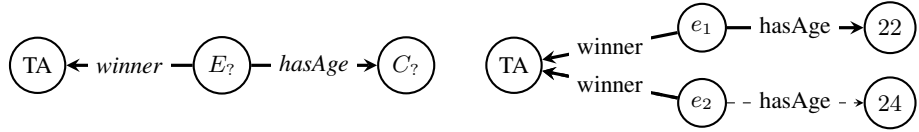

With this formulation, various questions can be asked on incomplete $\mathcal{G}_\mathcal{A}$.
For instance, the question ``\emph{What is the average age of Turing award (TA) winners?}'' can be answered by computing the mean of a set of numeric literals $C_?$:
\begin{equation}
\label{eq:turing_award_winners_avg_ge}
\small
\mean(C_?) .\exists E_?: \mathrm{winner}(E_?, \mathrm{turingAward}) \land \mathrm{hasAge}(E_?, C_? )
\end{equation}
Similarly, the question ``\emph{What is the minimum age of Turing award (TA) winners?}'' can be answered by computing the minimum of a set of numeric literals $C_?$:
\begin{equation}
\label{eq:turing_award_winners_min_age}
\small
\min(C_?) .\exists E_?: \mathrm{winner}(E_?, \mathrm{turingAward}) \land \mathrm{hasAge}(E_?, C_? )
\end{equation}
\Cref{fig:example-query-with-value-prediction} visualizes a subgraph of $\mathcal{G}_\mathcal{A}$ to answer $q$ defined in~\Cref{eq:turing_award_winners_avg_ge}. Having found the binding $S_?=\{e_1, e_2\}$ for $E_?$, to each $e \in S_?$, we apply the attribute predictor $\phi_\mathrm{winner}(e, \mathrm{turingAward})$ and average the results, yielding the answer $\llbracket q \rrbracket = \frac{22 + 24}{2}=23$---in contrast to $\llbracket q \rrbracket = 22$ by a symbolic approach that neglects missing information.

%



\section{Experimental Results}
\label{sec:experiments}


After a brief description of the experimental setup, we evaluate the performance of LitCQD on the query types shown in \Cref{tab:query_types}. Finally, we show the answers of LitCQD for an example query. Our code is publicly available.
\footnote{\url{https://github.com/dice-group/LitCQD}}





\begin{table}[tb]
    \caption{Different query types with their internal representation and how their answers are computed. Entity queries without literals were proposed by \citet{ren2020query2box}. Entity queries with literals and queries with literal answers are newly proposed in this paper.}
    \label{tab:query_types}
    \centering
    \footnotesize
    \setlength{\tabcolsep}{3pt}
        \begin{tabular}{@{}ll@{}}
            \toprule
            \multicolumn{2}{@{}c@{}}{\bfseries Multihop queries without literals} \\
            \midrule
            
            1p            & $E_?\:.\:r(e,E_?)$                                                                 \\

            2p            & $E_?\:.\:\exists E_1:r_1(e,E_1)\land r_2(E_1, E_?)$                                \\
            3p            & $E_?\:.\:\exists E_1E_2.r_1(e,E_1)\land r_2(E_1, E_2)\land r_3(E_2,E_?)$           \\
            2i            & $E_?\:.\:r_1(e_1,E_?)\land r_2(e_2,E_?)$                                           \\
            3i            & $E_?\:.\:r_1(e_1,E_?)\land r_2(e_2,E_?)\land r_3(e_3,E_?)$                         \\
            ip            & $E_?\:.\:\exists E_1.r_1(e_1,E_1)\land r_2(e_2,E_1)\land r_3(E_1,E_?)$             \\
            pi            & $E_?\:.\:\exists E_1.r_1(e_1,E_1)\land r_2(E_1,E_?)\land r_3(e_2,E_?)$             \\
            2u            & $E_?\:.\:r_1(e_1,E_?)\lor r_2(e_2,E_?)$                                            \\
            up            & $E_?\:.\:\exists E_1.[r_1(e_1,E_1)\lor r_2(e_2,E_1)]\land r_3(E_1,E_?)$            \\ \midrule
            \multicolumn{2}{@{}c@{}}{\bfseries Multihop queries with literals and entity answers} \\
            \midrule
            ai             & $E_?\:.\:\exists C_1.a(E_?,C_1)\land \mathit{af}(C_1,c)$                           \\
            2ai            & $E_?\:.\:\exists C_1C_2.a_1(E_?,C_1)\land \mathit{af_1}(C_1, c_1)\land a_2(E_?,C_2)\land \mathit{af_2}(C_2, c_2)$ \\
            pai            & $E_?\:.\:\exists V_1.r(e,E_?)\land a(E_?,C_1)\land \mathit{af}(C_1, c_1)$          \\
            aip            & $E_?\:.\:\exists E_1C_1.a(E_1,C_2)\land \mathit{af}(C_1, c_1)\land r(E_1,E_?)$     \\

            au             & $E_?\:.\:\exists C_1C_2.a_1(E_?,C_1)\land \mathit{af_1}(C_1, c_1)\lor a_2(E_?,C_2)\land \mathit{af_2}(C_2, c_2)$  \\
            \midrule
            \multicolumn{2}{@{}c@{}}{\bfseries Multihop queries with literals and literal answers} \\ 
            \midrule
            1ap            & $\mean(C_?)\:.\:a(e,C_?)$                                                          \\
            2ap            & $\mean(C_?)\:.\:\exists E_1.r(e,E_1)\land a(E_1,C_?)$                              \\
            3ap            & $\mean(C_?)\:.\:\exists E_1E_2.r_1(e,E_1)\land r_2(E_1, E_2)\land a(E_2,C_?)$      \\
            \bottomrule
        \end{tabular}
\end{table}

\begin{table*}[tb]
    \centering
    \setlength{\tabcolsep}{4pt}
    \small
    \caption{Query answering results with different attribute embedding models for multihop entity queries without literals. Results were computed for test queries over the FB15k-237 dataset and evaluated in terms of mean reciprocal rank (MRR) and Hits@k for $k\in\{1,3, 10\}$.
    }
    \label{tab:eval_queries_without_literals}
        \begin{tabular}{@{}lcccccccccc@{}}
            \toprule
            {\bf Method}            & {\bf Average}  & {\bf 1p}       & {\bf 2p}       & {\bf 3p}       & {\bf 2i}       & {\bf 3i}       & {\bf ip}       & {\bf pi}       & {\bf 2u}       & {\bf up}       \\
            \midrule
            \multicolumn{11}{c}{\bf MRR}                                                                                                                                                                      \\
            \midrule
            Query2Box               & 0.213          & 0.403          & 0.198          & 0.134          & 0.238          & 0.332          & 0.107          & 0.158          & 0.195          & 0.153          \\
            CQD & 0.295          & 0.454          & 0.275          & 0.197          & 0.339          & 0.457          & 0.188          & 0.267          & 0.261          & 0.214          \\
            LitCQD (ours) & \textbf{0.301} & \textbf{0.457} & \textbf{0.285} & \textbf{0.202} & \textbf{0.350} & \textbf{0.466}   & \textbf{0.193} & \textbf{0.274} & \textbf{0.266} & \textbf{0.215}          \\
            \midrule
            \multicolumn{11}{c}{\bfseries HITS@1}                                                                                                                                                                    \\
            \midrule
            Query2Box               & 0.124          & 0.293          & 0.120          & 0.071          & 0.124          & 0.202          & 0.056          & 0.083          & 0.094          & 0.079          \\
            CQD & 0.211          & 0.354          & 0.198          & 0.137          & 0.235          & 0.354          & 0.130          & 0.186          & 0.165          & \textbf{0.137} \\
            LitCQD (ours) & \textbf{0.215} & \textbf{0.355} & \textbf{0.206}   & \textbf{0.141} & \textbf{0.245} & \textbf{0.365}  & \textbf{0.129} & \textbf{0.193} & \textbf{0.168}   & 0.135          \\
            \midrule
            \multicolumn{11}{c}{\bfseries HITS@3}                                                                                                                                                                    \\
            \midrule
            Query2Box               & 0.240          & 0.453          & 0.214          & 0.142          & 0.277          & 0.399          & 0.111          & 0.176          & 0.226          & 0.161          \\
            CQD & 0.322          & 0.498          & 0.297          & 0.208          & 0.380          & 0.508          & 0.195          & 0.290          & 0.287          & \textbf{0.230}          \\
            LitCQD (ours) & \textbf{0.330} & \textbf{0.506} & \textbf{0.309} & \textbf{0.214} & \textbf{0.395} & \textbf{0.517} & \textbf{0.204} & \textbf{0.296}   & \textbf{0.295} & \textbf{0.235}          \\
            \midrule
            \multicolumn{11}{c}{\bfseries HITS@10}                                                                                                                                                                   \\
            \midrule
            Query2Box               & 0.390          & 0.623          & 0.356          & 0.259          & 0.472          & 0.580          & 0.203          & 0.303          & 0.405          & 0.303          \\
            CQD         & 0.463          & 0.656          & 0.422          & 0.312          & 0.551          & 0.656          & 0.305          & 0.425          & 0.465          & 0.370          \\
            LitCQD (ours)   & \textbf{0.472} & \textbf{0.660} & \textbf{0.439} & \textbf{0.323} & \textbf{0.561} & \textbf{0.663}  & \textbf{0.315} & \textbf{0.434} & \textbf{0.475} & \textbf{0.379}          \\
            \bottomrule
        \end{tabular}

\end{table*}
\begin{table*}[tb]
    \centering
    \setlength{\tabcolsep}{3pt}
    \small
    \caption{Query answering results for multihop entity queries with literals. Our best-performing model Complex-N3 + Attributes (KBLRN) is compared to variations thereof. Results were computed for test queries over the FB15k-237 dataset and evaluated in terms of Hit@10.}
    \label{tab:eval_query_results_attr_filtering}
        \begin{tabular}{@{}lccccccc@{}}
            \toprule
            \textbf{Method}                                 & \textbf{ai-lt} & \textbf{ai-eq} & \textbf{ai-gt} & \textbf{2ai} & \textbf{aip} & \textbf{pai} & \textbf{au} \\
            \midrule
            LitCQD   & 0.405   & 0.232          & 0.329          & 0.216        & 0.174        & 0.320 & 0.212       \\
            \midrule
            \quad - w/o attribute filter predictor  & 0.280  & 0.005     & 0.237          & 0.148      & 0.124        & 0.421   & 0.054  \\
            \quad - w/o attribute existence predictor   & 0.203         & 0.137          & 0.128    & 0.099        & 0.156        & 0.338     & 0.033       \\
            \quad - w/o both & 0.002          & 0.000          & 0.000          & 0.000        &0.086        & 0.412       & 0.002       \\
            \midrule
            \quad - w/o attribute-specific standard deviation    & 0.391  & 0.359         & 0.330   & 0.329    & 0.195     & 0.447      & 0.248    \\
            \bottomrule
        \end{tabular}

\end{table*}
\begin{table}[tb]
    \setlength{\tabcolsep}{5pt}
    \footnotesize
    \centering
    \caption{Query answering results for multihop literal queries for test queries over the FB15k-237 dataset evaluated in terms of mean absolute error (MAE) and  mean squared error (MSE).}
    \label{tab:eval_query_results_attr_pred}
        \begin{tabular}{@{}lcccccc@{}}
            \toprule
            \multirow{2}{*}{\textbf{Method}} & \multicolumn{2}{c}{\textbf{1ap}} & \multicolumn{2}{c}{\textbf{2ap}} & \multicolumn{2}{c}{\textbf{3ap}}                    \\
            \cmidrule(lr){2-3} \cmidrule(lr){4-5} \cmidrule(l){6-7}
                                            & MAE   & MSE   & MAE   & MSE   & MAE   & MSE   \\ \midrule
            LitCQD    & 0.050 & 0.011 & 0.034 & 0.005 & 0.041 & 0.007 \\ \midrule
            Mean Predictor  & 0.341 & 0.143 & 0.346 & 0.141 & 0.362 & 0.152 \\
            \bottomrule
        \end{tabular}

    \centering
\end{table}


%
\subsection{Experimental Setup}
\label{subsec:datasets}
\paragraph{Dataset and Query Generation}
We use the FB15k-237 dataset augmented with attributes as done by \citet{garcia2018kblrn}. The dataset contains 12,390 entities, 237 entity relations, 115 attribute relations, and 29,229 triples.
Queries and their expected answers are generated the same way as by~\citet{hamilton2018embedding}.
The newly introduced attribute filter conditions (af) are handled as follows: When checking for \emph{equality} ($\textit{af}(C,c)=\mathit{eq}(C, c)$), we consider all entities whose attribute value lies within one standard deviation from $c$ as correct where the standard deviation is computed per attribute relation $a$; when checking the \emph{less-than} or \emph{greater-than} criterion, the criterion is checked exactly, i.e., all entities with attribute value ``$\leq c$'' or ``$\geq c$'' are considered correct. \Cref{tab:query_types} gives an overview of the newly introduced query types along with previous query types.



\paragraph{Hyperparameters}

For each query type, we tried 16 different configurations on the validation set and chose the best before applying the model to the test set.
As our framework is derived from the CQD framework, it allows two different optimization algorithms: Continuous optimization (Co), Combinatorial optimization (Beam); two t-norms: Gödel (min), product (prod); and 7 different beam sizes $k\in \lbrace 2^2, 2^3, \dots, 2^8\rbrace$ for the combinatorial optimization algorithm.
Each optimization algorithm is computed for both of the t-norms resulting in 2 configurations using the continuous optimization algorithm and 14 using the combinatorial optimization algorithm as every beam size is evaluated for both t-norms.

\subsection{Multihop Queries without Literals}

In a first experiment (\Cref{tab:eval_queries_without_literals}), we compare the performance of our approach LitCQD to CQD~\cite{arakelyan2021complex} and Query2Box~\cite{ren2020query2box} on multihop entity queries without literals, which can be answered by all three models---in contrast to more expressive queries that can only be answered by LitCQD. While CQD does not utilize literal information and employs the vanilla ComplEx-N3~\cite{lacroix2018canonical} model, LitCQD employs a model combining ComplEx-N3~\cite{lacroix2018canonical} with TransEA~\cite{Wu2018TransEA}. \Cref{tab:eval_queries_without_literals} shows that LitCQD clearly outperforms CQD and Query2Box in terms of the mean reciprocal rank (MRR), and Hits@k for $k \in \{1,3,10\}$.

\subsection{Multihop Queries with Literals and Entity Answers}

\Cref{tab:eval_query_results_attr_filtering} shows the evaluation results for the new query types with filter restrictions introduced in \Cref{subsec:answering_entity_quers} (second block in \Cref{tab:query_types}). For the simple ai query, each filtering expression (\emph{less-than, equals, greater-than}) is evaluated separately; the other query types contain all three filtering expressions. Except for aip queries, all query types with literals can be answered with a performance of at least 0.329 which is comparable to query types without literals (cf. \Cref{tab:eval_queries_without_literals}).

Moreover, we experimented with different variants of our model and perform an ablation study. As described in \Cref{subsec:answering_entity_quers}, \Cref{eq:preliminary_attribute_filter_predictor}, the attribute filter predictor $\phi_{\textit{af}, a}$ is a product of $\phi_{\mathrm{exists},a}(e)$ and $\phi'_{\textit{af, a}}(\hat{c}, c)$. We performed three experiments, where we replaced each/both of the two scoring functions by the constant value 1. \Cref{tab:eval_query_results_attr_filtering} shows that both components are crucial and the performance drops drastically if one of them is removed.

Moreover, the \Cref{eq:attribute_filter_equal} and \Cref{eq:attribute_filter_lt} normalize the difference $\hat{c} - c$ by dividing by the standard deviation $\sigma_a$ that was computed on the set $C_a$ and thus depends on the attribute relation $a$. As an alternative, we computed a universal standard deviation across all attributes of the knowledge base, i.e., the standard deviation $\sigma$ of $\bigcup_{a \in \mathcal{A}} C_a$. \Cref{tab:eval_query_results_attr_filtering} (last line) shows that using a universal standard deviation instead of an attribute-specific standard deviation leads to a lower performance on 4 query types and to the same performance on the remaining 3 query types.

\subsection{Multihop Queries with Literals and Literal Answers}

\Cref{tab:eval_query_results_attr_pred} evaluates the performance of queries asking for literal answers. The predicted numeric values are compared to the actual numeric values in terms of mean absolute error (MAE) and mean squared error (MSE). Interestingly, we notice that the mean absolute error for the 2ap queries is lower than for 1ap queries. This can be explained by the fact that for 1ap queries a single prediction of an attribute value is made whereas 2ap queries average multiple predictions (the number of the beam width). For 3ap queries the performance drops again because the relation path becomes longer and errors accumulate.

As a simple baseline, we also report the results of the model that always predicts the mean value $\frac{1}{|\mathcal{C}_a|}\sum_{c \in \mathcal{C}_a} c$ of the attribute $a$ in the whole knowledge graph (mean predictor in the table).

\begin{table}[tb]
    \centering
    \small
    \caption{Ranking of LitCQD's top 10 answers to the query in \Cref{eq:example_query_usa} including their expected and predicted attribute value for \texttt{date\_of\_birth}. The star (*) indicates attribute values unseen during training and the double star (**) refers to attribute values not part of the dataset at all. The dash (--) indicates that an entity does not have a date of birth.
    }
    \setlength{\tabcolsep}{9pt}
    \label{tab:summary_attr_query}
        \begin{tabular}{@{}llcc@{}}
            \toprule
            Rank & {\bfseries Answer}     & {\bfseries Expected Attr.} & {\bfseries Predicted Attr.} \\ 
            \midrule
             1 & John Denver     & 1944,00                        & 1941,52                               \\ 
             2 & Donna Summer    & 1949,00                        & 1948,55                               \\ 
             3 & Rob Thomas      & 1972**                         & 1943,72                               \\ 
             4 & Funkadelic      & --                             & 1925,21                               \\ 
             5 & James Ingram    & 1952,17*                       & 1948,50                               \\ 
             6 & Dio             & 1942**                         & 1935,59                               \\ 
             7 & Spinal Trap     & --                             & 1942,65                               \\ 
             8 & Sheila E.       & 1958,00                        & 1960,93                               \\ 
             9 & Linus Pauling   & 1901,17                        & 1900,06                               \\ 
            10 & BT              & 1971,83*                       & 1955,80                               \\ 
            \bottomrule
        \end{tabular}
\end{table}

\subsection{Example Query and Answers}

As an illustration of the model's query-answering ability, consider the query ``What are musicians from the USA born before 1972?'' and its logical representation
\begin{equation}
\begin{split}
E_? \:.\: &\exists E_1 . \text{/music/artist/origin}(\text{USA},E_?) \land \\
&\text{/people/person/date\_of\_birth}(E_?, V_1) \land lt(E_1, 1972).
\end{split}
\label{eq:example_query_usa}
\end{equation}
\Cref{tab:summary_attr_query} lists the top 10 returned answers. Although the model confuses the bands \emph{Funkadelic} and \emph{Spinal Trap} as musicians with a date of birth, the model is able to produce a reasonable ranking of entities.
Out of these 10 entities, the entity \emph{Linus Pauling} receives the highest score of 0.95 for the attribute portion of the query. The model is confident that the entity has the attribute \texttt{/people/person/date\_of\_birth} and that its value is less than 1972.
The entity \emph{BT} only receives a score of 0.58 for the attribute portion of the query because its predicted value is closer to the threshold of 1972. The model is more certain that the connection \texttt{/music/artist/origin}, \emph{USA} exists for \emph{BT} compared to \emph{Linus Pauling}. Nevertheless, the learned embeddings implicitly encode that \emph{Linus Pauling} has another connection to the entity \emph{USA} via the \texttt{/people/person/nationality} relation. Hence, the model ranks \emph{Linus Pauling} before \emph{BT} when answering this query.


\section{Conclusion}
\label{sec:conclusion}

In this paper, we propose LitCQD, a novel approach to answer multihop queries on incomplete knowledge graphs with numeric literals. Our approach allows answering queries that could not be answered before, e.g., queries involving literal filter restrictions and queries predicting the value of numeric literals. Moreover, our experiments suggest that even the performance of answering multihop queries that could be answered before improves as the underlying knowledge graph embedding models now take literal information into account. This is an important finding as most real-world knowledge graphs contain millions of entities with numerical attributes.

In future work, we plan to further increase the expressiveness of our queries, e.g., by supporting string literals, Boolean literals as well as datetime literals.




\bibliographystyle{splncs04nat} 
\bibliography{references}

\begin{thebibliography}{34}
\providecommand{\natexlab}[1]{#1}
\providecommand{\url}[1]{\texttt{#1}}
\providecommand{\urlprefix}{URL }
\expandafter\ifx\csname urlstyle\endcsname\relax
  \providecommand{\doi}[1]{doi:\discretionary{}{}{}#1}\else
  \providecommand{\doi}{doi:\discretionary{}{}{}\begingroup
  \urlstyle{rm}\Url}\fi

\bibitem[{Adolphs et~al.(2011)Adolphs, Theobald, Sch{\"{a}}fer, Uszkoreit, and
  Weikum}]{adolphs2011YAGO-QA}
Adolphs, P., Theobald, M., Sch{\"{a}}fer, U., Uszkoreit, H., Weikum, G.:
  {YAGO-QA:} answering questions by structured knowledge queries. In: {ICSC},
  pp. 158--161, {IEEE} Computer Society (2011)

\bibitem[{Arakelyan et~al.(2021)Arakelyan, Daza, Minervini, and
  Cochez}]{arakelyan2021complex}
Arakelyan, E., Daza, D., Minervini, P., Cochez, M.: Complex query answering
  with neural link predictors. In: {ICLR}, OpenReview.net (2021)

\bibitem[{Auer et~al.(2007)Auer, Bizer, Kobilarov, Lehmann, Cyganiak, and
  Ives}]{auer2017DBpedia}
Auer, S., Bizer, C., Kobilarov, G., Lehmann, J., Cyganiak, R., Ives, Z.G.:
  Dbpedia: {A} nucleus for a web of open data. In: {ISWC/ASWC}, Lecture Notes
  in Computer Science, vol. 4825, pp. 722--735, Springer (2007)

\bibitem[{Balazevic et~al.(2019)Balazevic, Allen, and
  Hospedales}]{balavzevic2019tucker}
Balazevic, I., Allen, C., Hospedales, T.M.: Tucker: Tensor factorization for
  knowledge graph completion. In: {EMNLP/IJCNLP} {(1)}, pp. 5184--5193,
  Association for Computational Linguistics (2019)

\bibitem[{Bordes et~al.(2013)Bordes, Usunier, Garc{\'{\i}}a{-}Dur{\'{a}}n,
  Weston, and Yakhnenko}]{bordes2013translating}
Bordes, A., Usunier, N., Garc{\'{\i}}a{-}Dur{\'{a}}n, A., Weston, J.,
  Yakhnenko, O.: Translating embeddings for modeling multi-relational data. In:
  {NIPS}, pp. 2787--2795 (2013)

\bibitem[{Demir et~al.(2021)Demir, Moussallem, Heindorf, and
  Ngomo}]{demir2021hyperconvolutional}
Demir, C., Moussallem, D., Heindorf, S., Ngomo, A.N.: Convolutional
  hypercomplex embeddings for link prediction. In: {ACML}, Proceedings of
  Machine Learning Research, vol. 157, pp. 656--671, {PMLR} (2021)

\bibitem[{Demir and Ngomo(2021)}]{demir2021convolutional}
Demir, C., Ngomo, A.N.: Convolutional complex knowledge graph embeddings. In:
  {ESWC}, Lecture Notes in Computer Science, vol. 12731, pp. 409--424, Springer
  (2021)

\bibitem[{Dettmers et~al.(2018)Dettmers, Minervini, Stenetorp, and
  Riedel}]{dettmers2018convolutional}
Dettmers, T., Minervini, P., Stenetorp, P., Riedel, S.: Convolutional 2d
  knowledge graph embeddings. In: {AAAI}, pp. 1811--1818, {AAAI} Press (2018)

\bibitem[{Diefenbach et~al.(2017)Diefenbach, Tanon, Singh, and
  Maret}]{diefenbach2017question}
Diefenbach, D., Tanon, T.P., Singh, K.D., Maret, P.: Question answering
  benchmarks for wikidata. In: {ISWC} (Posters, Demos {\&} Industry Tracks),
  {CEUR} Workshop Proceedings, vol. 1963, CEUR-WS.org (2017)

\bibitem[{F{\"{a}}rber et~al.(2018)F{\"{a}}rber, Bartscherer, Menne, and
  Rettinger}]{Farber2018Linked}
F{\"{a}}rber, M., Bartscherer, F., Menne, C., Rettinger, A.: Linked data
  quality of dbpedia, freebase, opencyc, wikidata, and {YAGO}. Semantic Web
  \textbf{9}(1), 77--129 (2018)

\bibitem[{Gal{\'a}rraga et~al.(2015)Gal{\'a}rraga, Teflioudi, Hose, and
  Suchanek}]{galarraga2015fast}
Gal{\'a}rraga, L., Teflioudi, C., Hose, K., Suchanek, F.M.: Fast rule mining in
  ontological knowledge bases with amie++. The VLDB Journal \textbf{24}(6),
  707--730 (2015)

\bibitem[{Garc{\'{\i}}a{-}Dur{\'{a}}n and Niepert(2018)}]{garcia2018kblrn}
Garc{\'{\i}}a{-}Dur{\'{a}}n, A., Niepert, M.: Kblrn: End-to-end learning of
  knowledge base representations with latent, relational, and numerical
  features. In: {UAI}, pp. 372--381, {AUAI} Press (2018)

\bibitem[{Hamilton et~al.(2018)Hamilton, Bajaj, Zitnik, Jurafsky, and
  Leskovec}]{hamilton2018embedding}
Hamilton, W., Bajaj, P., Zitnik, M., Jurafsky, D., Leskovec, J.: Embedding
  logical queries on knowledge graphs. Advances in neural information
  processing systems \textbf{31} (2018)

\bibitem[{Heindorf et~al.(2022)Heindorf, Bl{\"{u}}baum, D{\"{u}}sterhus,
  Werner, Golani, Demir, and Ngomo}]{heindorf2022evolearner}
Heindorf, S., Bl{\"{u}}baum, L., D{\"{u}}sterhus, N., Werner, T., Golani, V.N.,
  Demir, C., Ngomo, A.N.: Evolearner: Learning description logics with
  evolutionary algorithms. In: {WWW}, pp. 818--828, {ACM} (2022)

\bibitem[{Klement et~al.(2004)Klement, Mesiar, and Pap}]{klement2004triangular}
Klement, E., Mesiar, R., Pap, E.: Triangular norms. position paper {I:} basic
  analytical and algebraic properties. Fuzzy Sets Syst. \textbf{143}(1), 5--26
  (2004)

\bibitem[{Kouagou et~al.(2022)Kouagou, Heindorf, Demir, and
  Ngomo}]{kouagou2022learning}
Kouagou, N.J., Heindorf, S., Demir, C., Ngomo, A.N.: Learning concept lengths
  accelerates concept learning in {ALC}. In: {ESWC}, Lecture Notes in Computer
  Science, vol. 13261, pp. 236--252, Springer (2022)

\bibitem[{Kristiadi et~al.(2019)Kristiadi, Khan, Lukovnikov, Lehmann, and
  Fischer}]{kristiadi2019incorporating}
Kristiadi, A., Khan, M.A., Lukovnikov, D., Lehmann, J., Fischer, A.:
  Incorporating literals into knowledge graph embeddings. In: {ISWC}, Lecture
  Notes in Computer Science, vol. 11778, pp. 347--363, Springer (2019)

\bibitem[{Lacroix et~al.(2018)Lacroix, Usunier, and
  Obozinski}]{lacroix2018canonical}
Lacroix, T., Usunier, N., Obozinski, G.: Canonical tensor decomposition for
  knowledge base completion. In: {ICML}, Proceedings of Machine Learning
  Research, vol.~80, pp. 2869--2878, {PMLR} (2018)

\bibitem[{Nickel et~al.(2016)Nickel, Murphy, Tresp, and
  Gabrilovich}]{nickel2016review}
Nickel, M., Murphy, K., Tresp, V., Gabrilovich, E.: A review of relational
  machine learning for knowledge graphs. Proc. {IEEE} \textbf{104}(1), 11--33
  (2016)

\bibitem[{Ren et~al.(2022)Ren, Dai, Dai, Chen, Zhou, Leskovec, and
  Schuurmans}]{ren2022smore}
Ren, H., Dai, H., Dai, B., Chen, X., Zhou, D., Leskovec, J., Schuurmans, D.:
  {SMORE:} knowledge graph completion and multi-hop reasoning in massive
  knowledge graphs. In: {KDD}, pp. 1472--1482, {ACM} (2022)

\bibitem[{Ren et~al.(2020)Ren, Hu, and Leskovec}]{ren2020query2box}
Ren, H., Hu, W., Leskovec, J.: Query2box: Reasoning over knowledge graphs in
  vector space using box embeddings. In: {ICLR}, OpenReview.net (2020)

\bibitem[{Ren and Leskovec(2020)}]{ren2020beta}
Ren, H., Leskovec, J.: Beta embeddings for multi-hop logical reasoning in
  knowledge graphs. In: NeurIPS (2020)

\bibitem[{da~Silva et~al.(2021)da~Silva, R{\"{o}}der, and
  Ngomo}]{silva2021using}
da~Silva, A.A.M., R{\"{o}}der, M., Ngomo, A.N.: Using compositional embeddings
  for fact checking. In: {ISWC}, Lecture Notes in Computer Science, vol. 12922,
  pp. 270--286, Springer (2021)

\bibitem[{Suchanek et~al.(2007)Suchanek, Kasneci, and
  Weikum}]{suchanek2007YAGO}
Suchanek, F.M., Kasneci, G., Weikum, G.: Yago: a core of semantic knowledge.
  In: {WWW}, pp. 697--706, {ACM} (2007)

\bibitem[{Sun et~al.(2019)Sun, Deng, Nie, and Tang}]{sun2019rotate}
Sun, Z., Deng, Z., Nie, J., Tang, J.: Rotate: Knowledge graph embedding by
  relational rotation in complex space. In: {ICLR} (Poster), OpenReview.net
  (2019)

\bibitem[{Tahri and Tibermacine(2013)}]{tahri2013dbpedia}
Tahri, A., Tibermacine, O.: Dbpedia based factoid question answering system.
  International Journal of Web \& Semantic Technology \textbf{4}(3), 23 (2013)

\bibitem[{Tay et~al.(2017)Tay, Tuan, Phan, and Hui}]{Tay2017MTKGNN}
Tay, Y., Tuan, L.A., Phan, M.C., Hui, S.C.: Multi-task neural network for
  non-discrete attribute prediction in knowledge graphs. In: {CIKM}, pp.
  1029--1038, {ACM} (2017)

\bibitem[{Trouillon et~al.(2016)Trouillon, Welbl, Riedel, Gaussier, and
  Bouchard}]{trouillon2016complex}
Trouillon, T., Welbl, J., Riedel, S., Gaussier, {\'{E}}., Bouchard, G.: Complex
  embeddings for simple link prediction. In: {ICML}, {JMLR} Workshop and
  Conference Proceedings, vol.~48, pp. 2071--2080, JMLR.org (2016)

\bibitem[{Vrandecic and Kr{\"{o}}tzsch(2014)}]{vrandecic2014Wikidata}
Vrandecic, D., Kr{\"{o}}tzsch, M.: Wikidata: a free collaborative
  knowledgebase. Commun. {ACM} \textbf{57}(10), 78--85 (2014)

\bibitem[{Wu and Wang(2018{\natexlab{a}})}]{Wu2018TransEA}
Wu, Y., Wang, Z.: Knowledge graph embedding with numeric attributes of
  entities. In: Rep4NLP@ACL, pp. 132--136, Association for Computational
  Linguistics (2018{\natexlab{a}})

\bibitem[{Wu and Wang(2018{\natexlab{b}})}]{wu2018knowledge}
Wu, Y., Wang, Z.: Knowledge graph embedding with numeric attributes of
  entities. In: Rep4NLP@ACL, pp. 132--136, Association for Computational
  Linguistics (2018{\natexlab{b}})

\bibitem[{Yang et~al.(2015)Yang, Yih, He, Gao, and Deng}]{yang2015embedding}
Yang, B., Yih, W., He, X., Gao, J., Deng, L.: Embedding entities and relations
  for learning and inference in knowledge bases. In: {ICLR} (Poster) (2015)

\bibitem[{Zhang et~al.(2019)Zhang, Tay, Yao, and Liu}]{zhang2019quaternion}
Zhang, S., Tay, Y., Yao, L., Liu, Q.: Quaternion knowledge graph embeddings.
  In: NeurIPS, pp. 2731--2741 (2019)

\bibitem[{Zhu et~al.(2022)Zhu, Galkin, Zhang, and Tang}]{zhu2022neural}
Zhu, Z., Galkin, M., Zhang, Z., Tang, J.: Neural-symbolic models for logical
  queries on knowledge graphs. In: {ICML}, Proceedings of Machine Learning
  Research, vol. 162, pp. 27454--27478, {PMLR} (2022)

\end{thebibliography}

\end{document}